\documentclass{article}
\usepackage{microtype}
\usepackage{graphicx}
\usepackage{subfigure}
\usepackage{booktabs} 

\usepackage{cleveref}
\usepackage{latexsym}
\usepackage{graphicx} 
\usepackage[export]{adjustbox}
\usepackage{algorithm}
\usepackage{algorithmic}
\usepackage{multirow}
\usepackage{mathrsfs}
\usepackage{bm}
\usepackage{amsmath}
\usepackage{amsfonts}
\usepackage{amssymb}
\usepackage{url}
\usepackage{amsmath}
\usepackage{enumitem}
\usepackage{blindtext}
\usepackage{verbatim}

\usepackage[accepted]{icml2018}
\icmltitlerunning{Equation Embeddings}
\begin{document}

\twocolumn[
\icmltitle{Equation Embeddings}



\icmlsetsymbol{equal}{*}

\begin{icmlauthorlist}
\icmlauthor{Kriste Krstovski}{cu}
\icmlauthor{David M. Blei}{cu}
\end{icmlauthorlist}

\icmlaffiliation{cu}{Columbia University, New York, NY 10027}
\icmlcorrespondingauthor{Kriste Krstovski}{kriste.krstovski@columbia.edu}
\icmlcorrespondingauthor{David M. Blei}{david.blei@columbia.edu}
\icmlkeywords{exponential family embeddings, equation embeddings}
\vskip 0.3in
]
\printAffiliationsAndNotice{} 

\begin{abstract}
  We present an unsupervised approach for discovering semantic
  representations of mathematical equations.  Equations are
  challenging to analyze because each is unique, or nearly unique. Our
  method, which we call \textit{equation embeddings}, finds good
  representations of equations by using the representations of their
  surrounding words. We used equation embeddings to analyze four
  collections of scientific articles from the arXiv, covering four
  computer science domains (NLP, IR, AI, and ML) and $\sim$98.5k
  equations.  Quantitatively, we found that equation embeddings
  provide better models when compared to existing word embedding
  approaches. Qualitatively, we found that equation embeddings provide
  coherent semantic representations of equations and can capture
  semantic similarity to other equations and to words.
\end{abstract}

\section{Introduction}
\label{sec:Intro}
Equations are an important part of scientific articles, but many
existing machine learning methods do not easily handle
them. They are challenging to work with because each is unique or
nearly unique; most equations occur only once.  An automatic understanding of equations,
however, would significantly benefit methods for analyzing
scientific literature. Useful representations of equations can help
draw connections between articles, improve retrieval of scientific
texts, and help create tools for exploring and
navigating scientific literature.

In this paper we propose equation embeddings (EqEmb), an unsupervised
approach for learning distributed representations of equations.  The
idea is to treat the equation as a "singleton word," one that appears
once but that appears in the context of other words. The surrounding
text of the equation---and in particular, the distributed
representations of that text---provides the data we need to develop a
useful representation of the equation.

\begin{figure*}[ht]
\includegraphics[width=\textwidth]{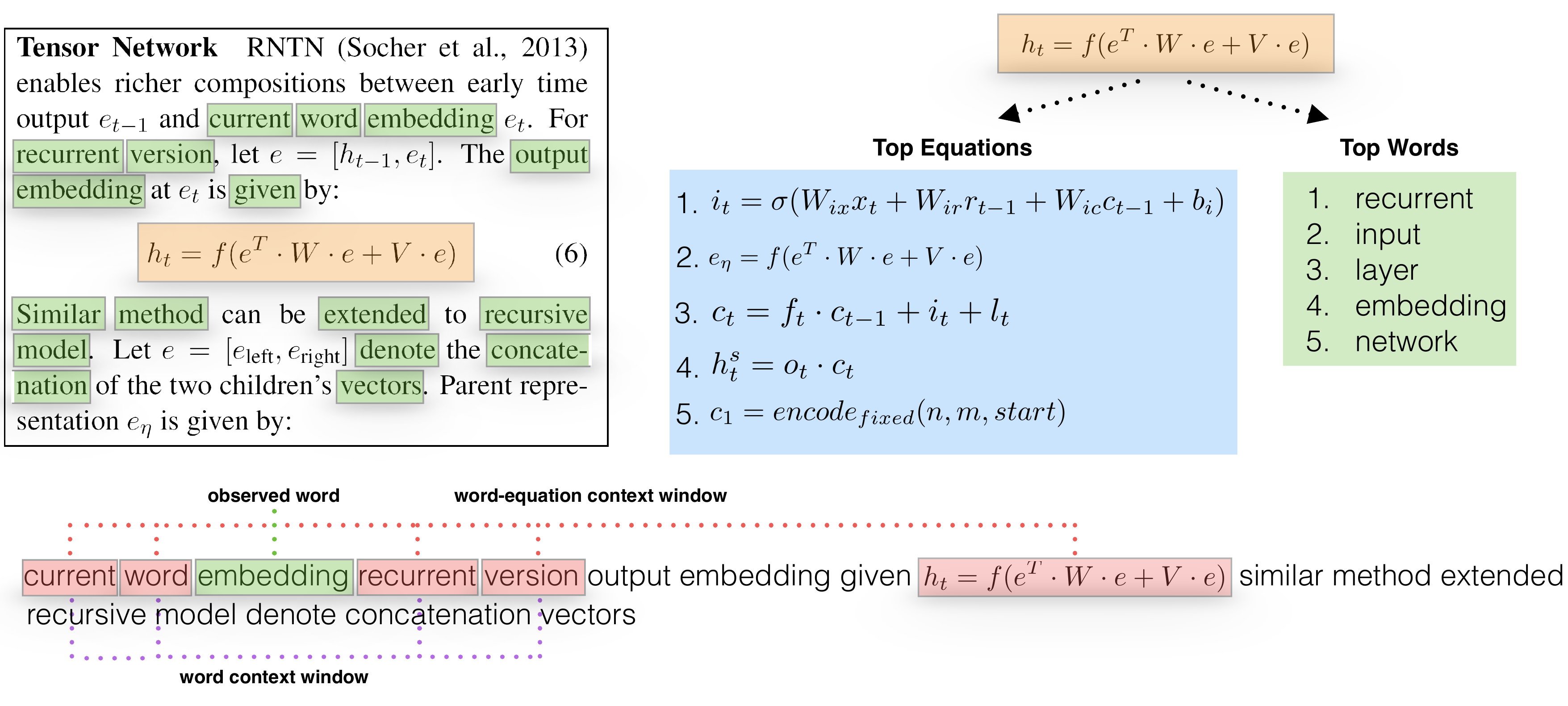}
\caption{\textit{Top left:} arXiv article snippet \cite{Li_ea_2015} that contains an equation of a neural network layer along with its surrounding context. Highlighted in green are words whose word-equation context window contains the equation. \textit{Top right:}  Extracted equation (top) and its 5 nearest equations (blue rectangle) and words (green rectangle) discovered using our approach. Discovered equations relate to neural network layers while nearest words bear semantic relatedness with the equation definition. \textit{Bottom:} Word-equation context window example. Highlighted in green is an example word whose word-equation context window contain the original context words (red) from the effective vocabulary that appear in a context window of size 4 along with the equation.}
\label{fig:nlp_example1}
\end{figure*}

Figure~\ref{fig:nlp_example1} illustrates our approach. On the left is an article snippet \cite{Li_ea_2015}. Highlighted in orange is an equation; in this example it represents a neural network layer. We note that this particular equation (in this form and with this notation) only occurs once in the collection of articles (from arXiv). The representations of the surrounding text, however, provide a meaningful context for the equation.  Those words allow us to learn its embedding, specifically as a "word" which appears in the context of its surroundings.  The resulting representation, when compared to other equations' representations and word representations, helps find both related equations and related words.  These are illustrated on the right.

EqEmbs build on exponential family embeddings~\cite{Rudolph_ea_2016}
to include equations as singleton observations and to model equation
elements such as variables, symbols and operators. Exponential family
embeddings, like all embedding methods, define a \textit{context} of
each word. In our initial EqEmb, the context for the words is a small
window, such as four or eight words, but the context of an equation is
a larger window, such as sixteen words. Using these two types of
contexts together finds meaningful representations of words and
equations. In the next EqEmb, which builds on the first, we consider
equations to be sentences consisting of equation units, i.e.,
variables, symbols, and operators. Equation units help model equations
across two types of context---over the surrounding units and over the
surrounding words.

We studied EqEmbs on four collections of scientific articles from the
arXiv, covering four computer science domains: natural language
processing (NLP), information retrieval (IR), artificial intelligence
(AI) and machine learning (ML).  We found that EqEmbs provide more
efficient modeling than existing word embedding methods. We further
carried out an exploratory analysis of a large set of $\sim$87k
equations. We found that EqEmbs provide better models when compared to
existing word embedding approaches. EqEmbs also provide coherent
semantic representations of equations and can capture semantic
similarity to other equations and to words.

\section{Related Work}
\label{sec:prev_work}
Word embeddings were first introduced in \citet{Rumelhart_ea_1986,
  Bengio_ea_2003} and there have been many variants
\cite{Mikolov_ea_2013a, Mikolov_ea_2013b, Pennington_ea_2014,
  Levy_Goldberg_2014}. Common for all of them is the idea that words
can be represented by latent feature vectors. These feature vectors
are optimized to maximize the conditional probability of the
dataset. Recently \citet{ Rudolph_ea_2016} extended the idea of word
embeddings to other types of data. EqEmb expand the idea of word
embeddings to a new type of data points -- equations.

There have been different proposed approaches for representing mathematical equations. \citet{Zanibbi_ea_2016} introduced the symbol layout tree, a representation that encodes the spatial relationship of variables and operators for the purpose of indexing and retrieving mathematical equations. Our work also falls into the framework of mathematical language processing (MLP) \cite{Lan_ea_2015} whose first step is converting mathematical solutions into a series of numerical features.
 
\section{Equation Embeddings Models}
\label{sec:eq_emb}

EqEmb are based on word embeddings \cite{Mikolov_ea_2013b} or specifically Bernoulli embeddings (b-embs) \cite{Rudolph_ea_2016}. Word embeddings models the probability of a word $w_i$ given its context $w_{c_{i}}$  as a conditional distribution $p(w_i | w_{c_{i}})$ where the context is defined as the set of words $w_j$ in a window of size $c_i$ that surrounds it. In word embeddings each word is assigned to two types of latent feature vectors, the embedding ($\rho$) and context ($\alpha$) vectors, both of which are $k$ dimensional. 

B-emb is an exponential family embedding model where the conditional distribution is a Bernoulli:
\begin{equation}
		p(w_i | w_{c_{i}}) = Bernoulli (b_w),
\label{eq:bernoulli}
\end{equation}

The parameter $b_w$ is defined using the word embedding $\rho_{w_i}$
and the word context $\alpha_{w_j}$ vectors:
\begin{equation}
		b_w = \sigma (\rho_{w_i}^T \sum_{j=1}^{| c_i |}{\alpha_{w_j}})
\label{eq:param}
\end{equation}
where $\sigma$ is the logistic function.
\subsection{Equation Embeddings}
Given a dataset of words and equations the goal of the EqEmb models is to derive a semantic representation of each equation. EqEmb model equations in the context of words. EqEmb is based on the idea that a good semantic representation of equations could be discovered by expanding the original word context to include any equations that appear in a possibly larger window around it. 

We assign embeddings to words ($\rho_{w}$, $\alpha_{w}$) and equations ($\rho_{e}$, $\alpha_{e}$). The objective function contains conditionals over the observed words and equations:
\begin{equation}
\begin{aligned}
\label{eq:eqemb_obj}
\begin{split}
\mathcal{L}(\rho_w,\alpha_w, \rho_e, \alpha_e)= \qquad \qquad \qquad \qquad \qquad \qquad \\
\sum_{i=1}^{W}\log(p(w_i | w_{c_{i}}))+
\sum_{m=1}^{E}\log(p(e_m | e_{c_{m}})).
\end{split}
\end{aligned}
\end{equation}

This is a sum of two sets of conditional distributions, the first over
observed words ($w_i$) and the second over observed equations
($e_m$). In word embedding models, $\rho$ and $\alpha$ are referred to
as embedding and context vectors.  Here we use a different
terminology: the interaction $\rho_e$ and feature vector $\alpha_e$.

In word embeddings, the context of the word $w_{c_{i}}$ is defined to index the surrounding words in a small window around it. Here the context of the word $w_i$ will be the original context ($\alpha_{w}$) and any equations ($\alpha_{e}$) that are in a possibly larger window around it. This is referred to as the \textit{word-equation context window}. 

Both conditionals are Bernoulli distributions. The first conditional is defined over the words in the collection. It has the following parameter:

\begin{eqnarray}
		b_w = \sigma(\rho_{w_i}^T g(\alpha_w,\alpha_e, c_i)).
\label{eq:objective_eta}
\end{eqnarray}

The word context function is:

\begin{eqnarray}
		g(\alpha_w,\alpha_e, c_i) = \sum_{j=1}^{| c_i |}\alpha_{w_{j}} + \sum_{k=1}^{| c'_i |}\alpha_{e_{k}}.
\label{eq:objective_g}
\end{eqnarray}

This function encompasses the words in the original word context ($\alpha_{w_j}$) and any equations ($\alpha_{e_k}$) that appear in a possibly larger window ($|c'_i|$) around it.  

The second term in the objective corresponds to the sum of the log conditional probabilities of each equation. Its  parameter is:
\begin{eqnarray}
		b_e = \sigma(\rho_{e_m}^T h(\alpha_w,c_m)).
\label{eq:objective_m}
\end{eqnarray}

Similar to word embeddings, equation context $e_{c_{m}}$ contains words that are in a context window around the equation:
\begin{eqnarray}
		h(\alpha_w, c_m) = \sum_{l=1}^{| c_m |}\alpha_{w_{l}}.
\label{eq:objective_h}
\end{eqnarray}

The equation context can have a larger window than the word context. Equation feature vectors ($\alpha_e$) are only associated with the first term of the objective function. This function contains the words where the equation appears in their larger context $c'_i$. 

The left side of Figure~\ref{fig:nlp_example1} shows an example
equation in a scientific article. With a word context of size $|c_i|$
we model the words in the article while ignoring equations. For
example when modeling the word "embedding" (highlighted in green) with
context window size of 4 (i.e. $|c_i|=4$), the context contains the
words that appear two words before ("current" and "word") and after
("recurrent" and "version") this word. With a word-equation context
window of size $|c'_i|$=16, the term for the word "embedding" would
have the feature vector of the equation as one of its components.

\subsection{Equation Unit Embeddings}
\begin{figure}[t]
\includegraphics[width=\columnwidth]{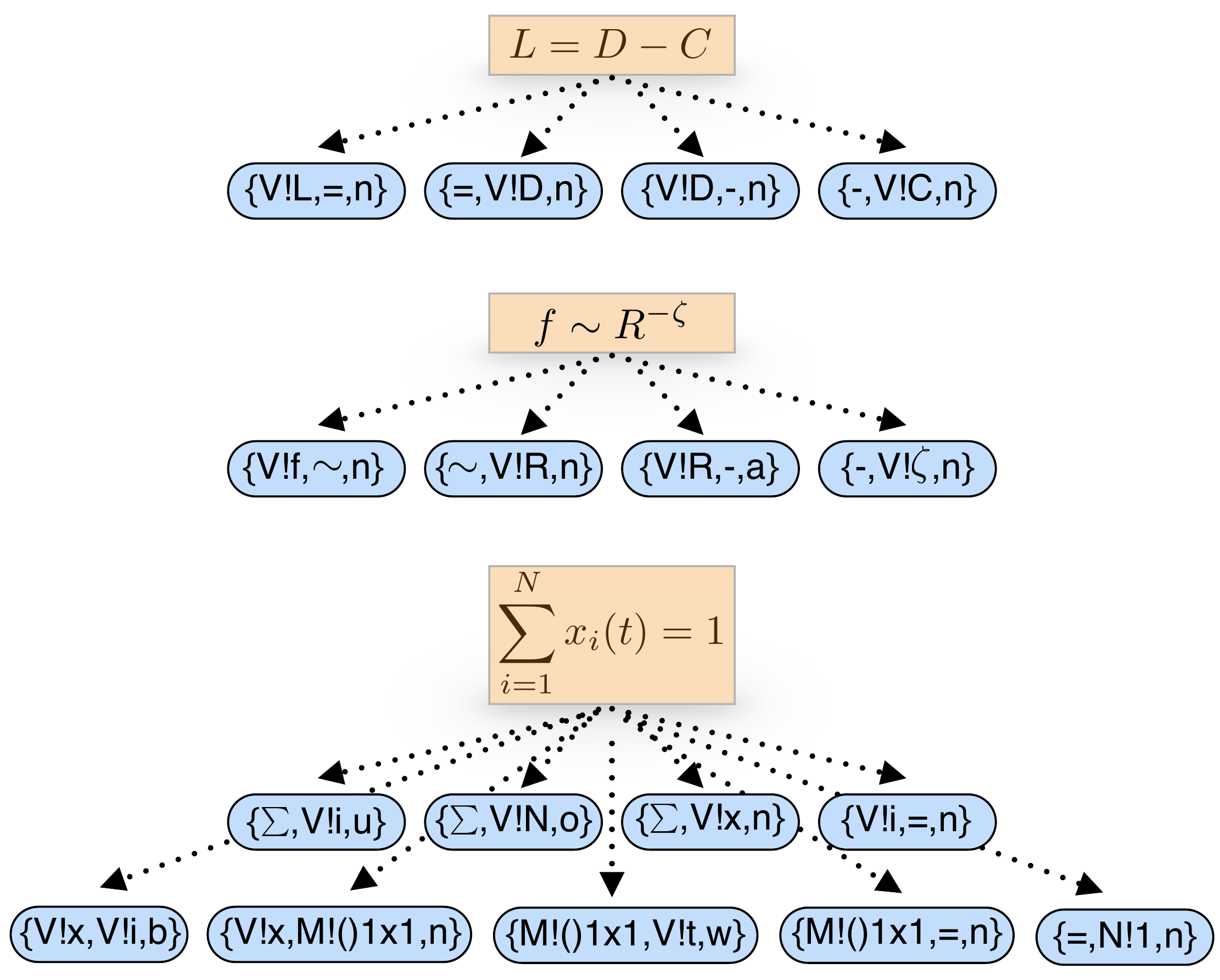}
\caption{Examples of Syntax Layout Tree (SLT) representation of equations using a symbol window of size one. Each tuple represents the special relationship between two symbols (\textit{n}-to the right; \textit{a}-above; \textit{u}-under; \textit{o}-over; \textit{w}-within).}
\label{fig:slt_example}
\end{figure}

Building on our previous method, we define a new model which we call
equation unit embeddings (EqEmb-U). EqEmb-U model equations by
treating them as sentences where the words are the equation variables,
symbols and operators which we refer to as units. The first step in
representing equations using equation units is to tokenize them. We
use the approach outlined in \citet{Zanibbi_ea_2016} which represents
equations into a syntax layout tree (SLT), a sequence of SLT tuples
each of which contains the spatial relationship information between
two equation symbols found within a particular window of equation
symbols. Figure~\ref{fig:slt_example} shows example SLT
representations of three equations.

Each equation $e$ is a sequence of equation units $u_j$, $j=1,2,...|e_{slt}|$ similar to a sentence where the words are the equation units. For each equation unit $u$ we assign interaction $\rho_{u}$ and feature $\alpha_{u}$ vectors. 

We assume that the context of the word $w_i$ will be the original context ($\alpha_w$) and the equation units ($\alpha_u$) of any equations that are in the word-equation context window. In addition for each equation unit we define its unit context $c_j$ to be the set of surrounding equation units in a small window $cs_u$ around it:
\begin{eqnarray}
		u_{c_j^u}=u_{j-cs_u/2},...,u_{j+cs_u/2}.
\label{eq:eq_unit_context}
\end{eqnarray}

The objective is over two conditionals, one for each context type:

\begin{equation}
\begin{aligned}
\label{eq:equ_objective}
\begin{split}
\mathcal{L}(\rho_w, \alpha_w, \rho_u, \alpha_u)= \qquad \qquad \qquad \qquad \qquad \\
\sum_{i=1}^{W}\log(p(w_i | w_{c_{i}}))+ 
\sum_{i=1}^{I}\log(p(u_{i} | u_{c_{i}})).
\end{split}
\end{aligned}
\end{equation}

The two parameters are:
\begin{eqnarray}
		b_w = \sigma(\rho_{w_i}^T (\sum_{j=1}^{| c_i |}\alpha_{w_{j}} + \sum_{k=1}^{| c'_i |} \sum_{l=1}^{| e_{k_{slt}}|}\alpha_{u_{k_l}})),\\
		b_u = \sigma(\rho_{u_{i}}^T \sum_{l=1}^{|c_l^u|}\alpha_{u_l}).
\label{eq:equ_objective_eta}
\end{eqnarray}

We define equation-level representations by averaging the representations of their constituent units:

\begin{eqnarray}
		\alpha_e = \frac{1}{|e_{slt}|} \sum_{j=1}^{|e_{slt}|}{\alpha_{u_j}}, \qquad \rho_e = \frac{1}{|e_{slt}|} \sum_{j=1}^{|e_{slt}|}{\rho_{u_j}}.
\label{eq:equ_objective_g}
\end{eqnarray}

\subsection{Computation}
We use stochastic gradient descent with Adagrad \cite{Duchi_ea_2011} to fit the embedding and context vectors.  Following \citet{Rudolph_ea_2016}, we reduce the computational complexity by splitting the gradient into two terms. The first term contains the non-zero entries ($x_i \neq 0$); the second term contains the zero entries ($x_i=0$). We compute the exact gradient for the non-zero points; We subsample for the zero data points. This is similar to negative sampling \cite{Mikolov_ea_2013b}, which also down-weights the contributions of the zero points. Unlike \citet{Rudolph_ea_2016} which uses $l_2$ regularization to protect against overfitting when fitting the embedding vectors we use early stopping based on validation accuracy, for the same effect. 

\section{Empirical Study}
\label{sec:emp_study}
We studied the performance of EqEmb on articles from the arXiv. EqEmb models provide better fits than
existing embedding approaches, and infer meaningful semantic relationships between equations and words in the collection.

We present a comparison of the proposed models to existing word embeddings approaches.
These are: the Bernoulli embeddings (b-emb) \cite{Rudolph_ea_2016}, continuous bag-of-words (CBOW)
\cite{Mikolov_ea_2013b}, Distributed Memory version of Paragraph Vector (PV-DM) \cite{Le_Mikolov_2014} and the Global Vectors (GloVe) \cite{Pennington_ea_2014} model.

\subsection{Datasets}
Our datasets are scientific articles that were published on arXiv. The sets contain articles (in LaTeX format) from four computer science domains: NLP, IR, AI, and ML. They were created by filtering arXiv articles based on their primary and secondary categories. We used the
following categories for the four collections: cs.cl for NLP; cs.ir
for IR; cs.ai for AI and stat.ml, stat.co, stat.me or cs.lg for ML. 

Table ~\ref{tab:coll_stats} shows the number of documents along with the number of unique words, equations and equation units for each collection. The equations are display equations that were enumerated in the LaTeX version of the articles. Unlike inline equations, which in many instances represent variables with general meaning (e.g. $x$, $y$, $\alpha$, etc.) and even numerical values, display equations typically represent mathematical concepts with more specific semantics. For the empirical study we used a random subset of 2k singletons from the total collection, along with all equations that occur more than once. For the qualitative analysis, we used all equations. 

We extracted words by tokenizing articles using the NLTK package \cite{Bird_ea_2009} and restricted the vocabulary to noun phrases and adjectives. The vocabulary was selected by:
\begin{itemize}
\item removing common stopwords
\item treating the top 25 most frequent words as stop words and removing them
\item including words whose term frequency is greater than or equal to 10 and whose character length is greater than or equal to 4
\item including the top 50 most frequent abbreviations whose character length is 3 (an exception to our previous rule)
\end{itemize}

When tokenizing equations, we first create an effective vocabulary of equation units. We convert equations into SLT format and collect collection wide frequency statistics over the equation units. The vocabulary contains all equation units whose frequency count is greater than $fc(u)\geq1$.

\begin{table}
\vskip 0.15in
\begin{center}
\begin{small}
\begin{sc}
\begin{tabular}{lrrrrr}
\toprule
Coll. &  \# Docs & \# Words   & \# Eqs. & \# Eq. Units \\
\hline
NLP      &     285  & 5,412     & 3,067  &  22,771 \\
IR       &     421  & 6,299     & 4,157  &  38,187 \\
AI       &   1,054  & 11,171   & 18,659  & 120,398 \\
ML       &   5,884  & 25,511   & 61,025  & 268,592 \\
\bottomrule
\hline
\end{tabular}
\end{sc}
\end{small}
\end{center}
\vskip -0.1in
\caption{Collections statistics across arXiv articles that we use in our analysis.}
\label{tab:coll_stats}
\end{table}

\begin{table*}
\centering
NLP Collection
\begin{center}
\begin{small}
\begin{sc}
\begin{tabular}{lrrrrrrrr}
\toprule
\hline
\multirow{3}{*}{Model} & \multicolumn{8}{c}{Latent Dimensions} \\
         &   \multicolumn{2}{c@{\quad}}{K=25} &  \multicolumn{2}{c@{\quad}}{K=50} &  \multicolumn{2}{c@{\quad}}{K=75} &  \multicolumn{2}{c@{\quad}}{K=100} \\
         & Validation &   Test  &  Validation &    Test & Validation &   Test  &  Validation &   Test  \\\cmidrule(r){2-3}\cmidrule(l){4-5}\cmidrule(r){6-7}\cmidrule(l){8-9}
CBOW     &    -11.52  &  -11.64 &     -11.24  &  -11.31 &    -11.56  &  -11.68 &     -11.56  &  -11.68 \\
PV-DM    &     -1.92  &   -1.93 &      -1.97  &   -1.97 &     -2.51  &   -2.51 &      -1.97  &   -1.95 \\
GloVe    &     -3.25  &   -3.21 &      -3.14  &   -3.10 &     -3.46  &   -3.40 &      -3.33  &   -3.27 \\
b-emb    &     -2.12  &   -2.12 &      -1.93  &   -1.96 &     -2.56  &   -2.51 &      -3.67  &   -3.75 \\
EqEmb    &     -1.51  &   -1.51 &      -1.51  &   -1.52 &     -1.64  &   -1.68 &      -1.97  &   -1.93 \\
EqEmb-U  &     \textbf{-1.47}  &   \textbf{-1.48} &      \textbf{-1.44}  &   \textbf{-1.45} &     \textbf{-1.52}  &   \textbf{-1.43} &      \textbf{-1.56}  &   \textbf{-1.61} \\
\hline
\bottomrule
\end{tabular}
 \end{sc}
\end{small}
\end{center}
\vspace*{.8cm}

\begin{center}
IR Collection
\end{center}
\begin{center}
\begin{small}
\begin{sc}
\begin{tabular}{lrrrrrrrr}
\toprule
\hline
\multirow{3}{*}{Model} & \multicolumn{8}{c}{Latent Dimensions} \\
         &   \multicolumn{2}{c@{\quad}}{K=25} &  \multicolumn{2}{c@{\quad}}{K=50} &  \multicolumn{2}{c@{\quad}}{K=75} &  \multicolumn{2}{c@{\quad}}{K=100} \\
         &        Validation &   Test  &  Validation &    Test & Validation &  Test  & Validation &   Test  \\\cmidrule(r){2-3}\cmidrule(l){4-5}\cmidrule(r){6-7}\cmidrule(l){8-9}
CBOW     &           -11.33  &  -11.22 &      -11.39 &  -11.31 &    -11.32  &  -11.2 &    -11.38  &  -11.29 \\
PV-DM    &            -2.29  &  -2.27  &      -2.29  &   -2.26 &     -2.31  &  -2.27 &     -2.34  &   -2.31 \\  
GloVe    &            -4.19  &  -4.09  &      -1.88  &   -1.83 &     -2.68  &  -2.61 &     -4.16  &   -4.04 \\
b-emb    &            -1.60  &  -1.61  &      -1.82  &   -1.80 &     -2.20  &  -2.22 &     -2.19  &   -2.28 \\
EqEmb    &            -1.60  &  -1.58  &      \textbf{-1.51}  &   \textbf{-1.52} &     -1.41  &  -1.44 &     \textbf{-1.41}  &   \textbf{-1.43} \\
EqEmb-U  &            \textbf{-1.21}  &  \textbf{-1.20}  &      -1.58  &   -1.57 &     \textbf{-1.14}  &  \textbf{-1.11} &     -1.47  &   -1.51 \\
\hline
\bottomrule
\end{tabular}
 \end{sc}
\end{small}
\end{center}

\vspace*{.8cm}

\begin{center}
AI Collection
\end{center}
\begin{center}
 \begin{small}
\begin{sc}
\begin{tabular}{lrrrrrrrr}
\toprule
\hline
\multirow{3}{*}{Model} & \multicolumn{8}{c}{Latent Dimensions} \\
         &   \multicolumn{2}{c@{\quad}}{K=25} &  \multicolumn{2}{c@{\quad}}{K=50} &  \multicolumn{2}{c@{\quad}}{K=75} &  \multicolumn{2}{c@{\quad}}{K=100} \\
         & Validation &   Test  &  Validation &    Test & Validation &   Test  &  Validation &   Test  \\\cmidrule(r){2-3}\cmidrule(l){4-5}\cmidrule(r){6-7}\cmidrule(l){8-9}
CBOW     &    -10.06  &  -10.19 &     -10.03  &  -10.03 &  -9.95  &  -9.99 &  -9.98  &  -10.11 \\
PV-DM    &     -3.48  &   -3.59 &      -3.63  &   -3.69 &  -3.56  &  -3.68 &  -3.69  &   -3.80 \\ 
GloVe    &     -1.47  &   -1.48 &       1.60  &   -1.58 &  -2.39  &  -2.43 &  -2.97  &   -3.01 \\
b-emb    &     -2.08  &   -2.05 &      -2.53  &   -2.52 &  -2.62  &  -2.47 &  -2.72  &   -2.67 \\
EqEmb    &     -1.38  &   -1.36 &      -1.39  &   -1.38 &  -1.45  &  -1.43 &  -1.52  &   -1.49 \\
EqEmb-U  &     \textbf{-1.37}  &   \textbf{-1.35} &      \textbf{-1.28}  &   \textbf{-1.27} &  \textbf{-1.44}  &  \textbf{-1.42} &  \textbf{-1.41}  &   \textbf{-1.41} \\
\hline
\bottomrule
\end{tabular}
\end{sc}
\end{small}
\end{center}

\vspace*{.8cm}
\begin{center}
ML Collection
\end{center}
\begin{center}
\begin{small}
\begin{sc}
\begin{tabular}{lrrrrrrrr}
\toprule
\hline
\multirow{3}{*}{Model} & \multicolumn{8}{c}{Latent Dimensions} \\
         &   \multicolumn{2}{c@{\quad}}{K=25} &  \multicolumn{2}{c@{\quad}}{K=50} &  \multicolumn{2}{c@{\quad}}{K=75} &  \multicolumn{2}{c@{\quad}}{K=100} \\
         & Validation &   Test  &  Validation &    Test & Validation &   Test  &  Validation &   Test  \\\cmidrule(r){2-3}\cmidrule(l){4-5}\cmidrule(r){6-7}\cmidrule(l){8-9}
CBOW     &    -11.26  &  -11.25 &     -11.22  &  -11.15 &    -11.33  &  -11.23 &     -11.27  &  -11.2 \\
PV-DM    &     -2.86  &   -2.88 &      -2.87  &   -2.88 &     -2.22  &   -2.23 &      -2.22  &  -2.26 \\
GloVe    &     -4.03  &   -4.11 &      -4.00  &   -4.07 &     -3.95  &   -4.02 &      -3.41  &  -3.46 \\
b-emb    &     -1.83  &   -1.82 &      -1.91  &    -1.9 &     -2.56  &   -2.44 &      -2.55  &  -2.71 \\
EqEmb    &     -1.53  &   -1.52 &      -1.57  &   -1.58 &     -1.75  &   -1.74 &      -1.92  &  -1.95 \\
EqEmb-U  &     \textbf{-1.42}  &   \textbf{-1.43} &      \textbf{-1.45}  &   \textbf{-1.46} &     \textbf{-1.62}  &   \textbf{-1.64} &      \textbf{-1.71}  &  \textbf{-1.74} \\
\hline
\bottomrule
\end{tabular}
 \end{sc}
\end{small}
 \end{center}
\caption{EqEmb outperform previous embedding models; EqEmb-U further improves performance. Performance comparisons between CBOW, GloVe, PV-DM, b-emb, EqEmb and EqEmb-U using log-likelihood computed on test and validation datasets. Comparisons were done over 4 different collections of scientific articles (NLP, IR, AI and ML) and across different latent dimensions (K=25, 50, 75 and 100).}
\label{tab:comp_res}
\end{table*}

\begin{table*}
\centering
{\renewcommand{\arraystretch}{1.8}
\begin{tabular}{ll}

\toprule
\hline
\multicolumn{2}{c}{$\cos(\textbf{x},\textbf{y}) = \frac{\sum^n_{i=1}x_{i} \cdot y_{i}}{\sqrt{\sum^n_{i=1}x^2_{i} \cdot \sum^n_{i=1} y^2_{i}}}$}\\
\midrule
\textbf{Rank.} & \multicolumn{1}{c}{\textbf{Top Equations}} \\
\hline
1.   &    $\cos ({\bf t},{\bf e})= {{\bf t} {\bf e} \over \|{\bf t}\| \|{\bf e}\|} = \frac{ \sum_{i=1}^{n}{{\bf t}_i{\bf e}_i} }{ \sqrt{\sum_{i=1}^{n}{({\bf t}_i)^2}} \sqrt{\sum_{i=1}^{n}{({\bf e}_i)^2}} }$\\
2.   &   $\rho(x,y) = \frac{\displaystyle\sum_{i = 1}^n(x_{i} - x_{\mu})(y_{i} - y_{\mu})}{\sqrt{\displaystyle\sum_{i = 1}^n(x_{i} - x_{\mu})^{2}\sum_{i = 1}^n(y_{i} - y_{\mu})^{2}}},$ \\
3.   &    $\mathrm{Sim}_{\gamma} (P, Q) = \frac{\sum_{i = 1}^\ell p_i^\gamma q_i^\gamma} {\sqrt{\sum_{i = 1}^\ell p_i^{2\gamma}} \sqrt{\sum_{i = 1}^\ell q_i^{2\gamma}}}$\\
4.   &   $similarity(v_p, v) = \frac{\sum_i^n v_{p_i}  v_i}{\sqrt{\sum_i^n v_{p_i}^2}  \sqrt{\vphantom{\sum_i^n v_{p_i}^2} \sum_i^n v_i^2}}$ \\
5.   &   $\dot x_i = \rho x_i \left( \sum_{j=1}^N x_j u(x_j) -u(x_i) \right)+\frac{\mu}{L} \left( \sum_{j=1}^N w_{ij} x_j -x_i \sum_{j=1}^N w_{ji} \right).$ \\
\hline
\bottomrule
\end{tabular}}
\caption{Example query equation (top row) and its 5 nearest equations discovered using EqEmb-U.}
\label{tab:cos_eq_example}
\vspace*{0.8cm}

{\renewcommand{\arraystretch}{1.6}
\begin{tabular}{clcl}
\toprule
\hline
\multicolumn{4}{c}{  $p(w|d)=\sum_{t=1}^{T}{p(w|t)p(t|d)}+\sum_{c=1}^{C}{p(w|c)p(T+c|d)}$}\\
\midrule
\multicolumn{2}{c@{\quad}}{\textbf{Top Equations}} & \multicolumn{2}{c@{\quad}}{\textbf{Top Words}} \\\cmidrule(r){1-2}\cmidrule(l){3-4}
1.  & $p(x_i=0,z_i=t|w_i=w,w_{-i},x_{-i},z_{-i},\gamma,\alpha,\tau,\beta_{\phi},\beta_{\psi}) $  &1.  & lda \\
2.  &  $\alpha_t^{new}=\alpha_t^{old}=\frac{\sum_d{\Psi(C_{td}+\alpha_t)-\Psi (\alpha_t))}}{\sum_d{(\Psi(\sum_{t'}{C_{t'd}}+\sum_{t'}{\alpha_t^{'}}-\Psi(\sum_{t^{'}}{\alpha_t'}))}}$   & 2.  & topics \\
3.  &  $p(v,y)=p(y|v)\prod_{i=1}^{D_v}{p(v_i|v_{<i})}$   &3.  & concept \\
4.  &  $p(\overrightarrow{n})=Mult(\overrightarrow{n}|\overrightarrow{p},N)= {{N}\choose{\overrightarrow{n}}} \ \prod_{k=1}^{V}{p_k^{n_k}}$   & 4.  & concept-topic \\
5.  &  $ p(\beta_{1:K},\theta_{1:D},z_{1:D}|w_{1:D})=\frac{p(\beta_{1:K},\theta_{1:D},z_{1:D},w_{1:D})}{p(w_{1:D})}$ & 5.  & concepts \\
\hline
\bottomrule
\end{tabular}}
\caption{Example query equation (top row) and its 5 nearest equations (left) and words (right) discovered using EqEmb. All similar equations relate to the LDA model.}
\label{tab:nlp_example2}
\vspace*{0.8cm}
{\renewcommand{\arraystretch}{1.6}
\begin{tabular}{clcl}
\toprule
\hline
\multicolumn{4}{c}{  $F\textnormal{-}measure=\frac{2PrecisionRecall}{Precision+Recall}$} \\
\midrule
\multicolumn{2}{c@{\quad}}{\textbf{Top Equations}} & \multicolumn{2}{c@{\quad}}{\textbf{Top Words}} \\\cmidrule(r){1-2}\cmidrule(l){3-4}
1.  & $Recall=\frac{Number\ of\ matched\ frames(color\ and\ texture)}{Number\ of\ frames\ in\ user\ summary}$  &1.  & f-measure \\
2.  &  $ROUGE-N-P = \frac{\sum_{I \in CT}{\sum_{gram_N \in I}{C_{nt_{match}}(gram_N)}}}{\sum_{I \in CT}{\sum_{gram_N \in I}{C_{n{t}}(gram_N)}}}$   & 2.  & test \\
3.  &  $precision=\frac{n_{correct}}{n_{total}}$   &3.  & report \\
4.  &  $P@k=\frac{number\ of\ actual\ friends\ at\ top\ k}{k}$   & 4.  & performance \\
5.  &  $FNR=\frac{FN}{FN+TP},\ \ \ FPR=\frac{FP}{FP+TP}$ & 5.  & accuracy \\
\hline
\bottomrule
\end{tabular}}
\caption{Example query equation (top row) and its 5 nearest equations (left column) and words (right column) discovered using EqEmb. All similar equations relate to classification performance measures such as F-measure. }
\label{tab:ir_example1}
\end{table*}

\begin{table}[h]
{\renewcommand{\arraystretch}{1.5}
\begin{tabular}{p{0.7cm}rrrr}
\toprule
\hline
\multicolumn{5}{c}{ $\cos(\textbf{x},\textbf{y}) = \frac{\sum^n_{i=1}x_{i} \cdot y_{i}}{\sqrt{\sum^n_{i=1}x^2_{i} \cdot \sum^n_{i=1} y^2_{i}}}$}\\
\midrule
Rank  & CBOW     & PV-DM          & GloVe         & EqEmb       \\
\hline
1.    & vectors & vectors         & accepted      & cosine      \\
2.    & inner   & inner           & brothers      & distance    \\
3.    & space   & vector          & partition     & similarity  \\
4.    & center  & angle           & stimulus      & metric      \\
5.    & good    & dense           & miss          & cluster     \\
\hline
\bottomrule
\end{tabular}}
\caption{Example query equation (top row) and its five most similar words obtained using CBOW, PV-DM, GloVe and EqEmb.}
\label{tab:eq_word_sim}
\vspace*{1cm}
{\renewcommand{\arraystretch}{1.7}
\begin{tabular}{p{0.7cm}l}
\toprule
\hline
\multicolumn{2}{c}{\textbf{Query:} similarity, distance, cosine}\\
\midrule
\textbf{Rank.} &  \multicolumn{1}{c}{\textbf{Top Equations}} \\
\hline
1.   & $\cos(\textbf{x},\textbf{y}) = \frac{\sum^n_{i=1}x_{i} \cdot y_{i}}{\sqrt{\sum^n_{i=1}x^2_{i} \cdot \sum^n_{i=1} y^2_{i}}}$      \\
2.   & $\mathrm{Sim}_{\alpha} (P, Q) = \frac{ \sum_1^\ell p_i^\alpha q_i^\alpha} {\sum_1^\ell( p_i^{2\alpha} + q_i^{2\alpha} - p_i^\alpha q_i^\alpha) }$      \\
3.   & $\mathrm{Sim}_{\gamma} (P,Q) = \frac{\sum_{i = 1}^\ell p_i^\gamma q_i^\gamma} {\sqrt{\sum_{i = 1}^\ell p_i^{2\gamma}} \sqrt{\sum_{i = 1}^\ell q_i^{2\gamma}}}$      \\
4.   & $\mathrm{Dist}_k (D_1,D_2) \equiv 1 - \mathrm{Res}_k(D_1, D_2). $      \\
5.   & $dist(x,y) = ||x-y||_2^2 = (\sum^{m}_{i=1}(x_i - y_i)^2)^{(1/2)}$      \\
\hline
\bottomrule
\end{tabular}}
\caption{Example query which consists of 3 words ("similarity", "distance" and "cosine") and its 5 nearest equations discovered using EqEmb. For a given word query, EqEmb are able to retrieve query relevant equations. }
\label{tab:query_eq_sim}
\end{table}

\subsection{Experimental Setup}
We analyzed EqEmb models performance using a held out set of words that we generate for each equation in our collections. Held out sets are constructed using the following procedure: We traverse over the collections and for every discovered equation we randomly sample words from its context set. The held out set contains the sampled words and their context window which also includes the equation. For each held out word we also generate a set of negative samples for the given word context. We perform the same procedure to form a validation set. For each of the $l$ equations in a collection, two held out words $w_h^e$ are sampled. For a context window of size 4 the sampled word context is defined as $w_{c}^e=[w_{c_1},w_{c_2},w_{c_3},e]$. 

During training we compute the predictive log-likelihood on the validation set of words using the fitted model after each iteration over the collection. A fitted model is a collection of interaction and feature vectors for each equation and word. Given a fitted model, the log probability of a held out word is computed using the following formula: 
\begin{equation}
p(w_h^e|w_{c}^e)= \frac{\exp({\rho(w_h^e)}^T \sum_{j=1}^{|w_{c}^e|}\alpha(w_{{c}_j}^e))}{\sum_{k=1}^{e^{|ns|}+1} {\exp({\rho(w_k)}^T \sum_{j=1}^{|c|}\alpha(w_{{c}_j}^e))}}
\end{equation}
which is the softmax function computed over a set of negative samples $|ns|$ and the held out word. In particular, we ran the model 20 times across the collection. After each collection iteration $i$ we observe whether the predictive log-likelihood continues to improve compared to the previous ($i-1$) iteration. We stop at the $i$-th iteration when that is no longer the case.

When modeling equations using EqEmb we perform two passes over the collection. In the first pass we only model words while ignoring equations. In the second pass we only model equations while holding fixed the interaction and feature vectors of all words. In context of EqEmb we treat equations as singleton words and the broader question that we are trying to answer is whether we can learn something about the meaning of the singleton words given the fixed word interaction and feature vectors.

\subsubsection{Evaluation Models and Parameters}
In our analysis we evaluated the performance of the EqEmb models across different sizes for the word context (W), word-equation context (E) and embedding vector size (K). Model performance was compared with 4 existing embedding models: b-emb\footnote{\label{efemb}\url{https://github.com/mariru/exponential_family_embeddings}}, CBOW, GloVe\footnote{\url{https://nlp.stanford.edu/projects/glove}} and PV-DM. We used the gensim \cite{Rehurek_Stojka_2010} implementation of the CBOW and PV-DM models. When modeling equations using the first 3 embedding models we treat equations as regular words in the collection. In case of the PV-DM model we parse the article so that equations and their surrounding context of length equivalent to the word-equation context window are labeled as a separate paragraph. We also assign paragraph labels to the article text occurring between equation paragraphs.  

\subsection{Results}
Table ~\ref{tab:comp_res} shows the performance comparison results across the different embeddings models. For each model, performance results are shown on 4 latent dimension values (K=25, 50, 75 and 100). For each dimension we ran experiments by varying the context window size for words (Word Context=4, 8 and 16). In addition for the EqEmb, EqEmb-U and PV-DM models we also varied the word-equation window size (E=8 and 16). Comparisons across models are performed using the pseudo log-likelihood measure \cite{Rudolph_ea_2017}. For a given held-out word $w_h^e$ and a set of negative samples $w_{ns}$ the pseudo log-likelihood is defined as: 
\begin{equation}
\log(p(w_h^e|w_{c}^e))+ \frac{1}{|w_{ns}^e|}\sum_{j=1}^{|w_{ns}^e|}{log(1- p(w_{{ns}_{j}}|w_{c}^e))}. 
\end{equation}
We treat this a 
downstream task. For each model type and latent dimension configuration, we use the validation set to select the best model configuration (i.e. combination of context window sizes). We report values on both datasets.

Across all collections EqEmb outperform previous embedding models and EqEmb-U further improves performance. 

\subsection{Word Representation of Equations}
\label{sec:word_rep}
EqEmb help obtain word descriptions of the equations. Table ~\ref{tab:eq_word_sim} shows example equation and the 5 most similar words obtained using 4 different embedding approaches which include CBOW, PV-DM, GloVe and EqEmb. For the query equation we obtain most similar words by computing Cosine distance between the embedding vector ($\rho_e$) representation of the query equation and the context vector representation of the words ($\alpha_w$). 

With the embedding representation of words and equations we could also perform equation search using words as queries. For a set of query words we generate its embedding representation by taking the average of the embedding representation of each word and compute Cosine distance across all the equations embeddings. Table ~\ref{tab:query_eq_sim} shows an example query, which consists of three words, and its 5 nearest equations discovered using EqEmb. For a given word query, EqEmb are able to retrieve query relevant equations. 

\subsection{Discovering Semantically Similar Equations}
\label{sec:semsim_eq}

In addition to words, EqEmb models can capture the semantic similarity between equations in the collection. We performed qualitative  analysis of the model performance using all discovered equations across the 4 collection. Table~\ref{tab:cos_eq_example} shows the query equation used in the previous analysis and its 5 most similar equations discovered using EqEmb-U. For qualitative comparisons across the other embedding models, in Appendix A we provide results over the same query using CBOW, PV-DM, GloVe and EqEmb.  In Appendix A reader should notice the difference in performance between EqEmb-U and EqEmb compared to existing embedding models which fail to discover semantically similar equations. \Cref{tab:ir_example1,tab:nlp_example2} show two additional example equation and its 5 most similar equations and words discovered using the EqEmb model. Similar words were ranked by computing Cosine distance between the embedding vector ($\rho_e$) representation of the query equation and the context vector representation of the words ($\alpha_w$). Similar equations were discovered using Euclidean distance computed between the context vector representations of the equations ($\alpha_e$). We give additional example results in Appendix B.

\section{Conclusion}
\label{sec:con}
We presented unsupervised approaches for semantic representations of mathematical equations using their surrounding words. Across 4 different collections we showed that out methods offer more effective modeling compared to existing embedding models. We also demonstrate that they can capture the semantic similarity between equations and the words in the collection. In the future we plan to explore how EqEmb could be expend to represent other objects such as images, captions and inline figures.

\bibliography{eqemb}
\bibliographystyle{icml2018}

\onecolumn
\appendix
\section{Equation Similarity Results Obtained Using CBOW, PV-DM, GloVe and EqEmb}
\label{sec:cosine_example}
Shown in the following 4 tables are the top five equations obtain for the example query using CBOW, PV-DM, GloVe and EqEmb. 

\begin{table}[h]
\begin{minipage}{\textwidth}
\centering
\begin{center}
\begin{small}
{\renewcommand{\arraystretch}{1.4}
\begin{tabular}{ll}
\toprule
\hline
\multicolumn{2}{c}{Model: CBOW} \\
\midrule
\multicolumn{2}{c}{Query Equation: $\cos(\textbf{x},\textbf{y}) = \frac{\sum^n_{i=1}x_{i} \cdot y_{i}}{\sqrt{\sum^n_{i=1}x^2_{i} \cdot \sum^n_{i=1} y^2_{i}}}$}\\
\midrule
Rank. &  \multicolumn{1}{c}{\textbf{Top Equations}} \\
\hline
1.   &   $Sw_1 = \lambda_{1}w_{1}.$\\
2.   &  $\phi_j ={1 \over f_j} u_j$ \\
3.   &  $\sum_j {f_{ij} \over f_i} \phi_j$\\
4.   &  $x_j^* = \frac{x_{j - \lfloor n/2 \rfloor} + ... + x_{j-1} + x_j + x_{j+1} + ... + x_{j + \lfloor n/2 \rfloor} }{n},$ \\
5.   &  $rank(s,t) = alignmentScore(s,t) * languageModelScore(t)$ \\
\hline
\bottomrule
\end{tabular}}
\end{small}
\end{center}
\vskip -0.1in
\caption{CBOW: Example query equation (top row) and its top 5 nearest equations discovered with this model.}
\end{minipage}\hfill

\vspace*{0.8cm}

\begin{minipage}{\textwidth}
\centering
\begin{center}
\begin{small}
{\renewcommand{\arraystretch}{1.4}
\begin{tabular}[b]{ll}
\toprule
\hline
\multicolumn{2}{c}{Model: PV-DM} \\
\midrule
\multicolumn{2}{c}{Query Equation: $\cos(\textbf{x},\textbf{y}) = \frac{\sum^n_{i=1}x_{i} \cdot y_{i}}{\sqrt{\sum^n_{i=1}x^2_{i} \cdot \sum^n_{i=1} y^2_{i}}}$}\\
\midrule
Rank. &  \multicolumn{1}{c}{\textbf{Top Equations}} \\
\hline
1.   &  $\Re\langle A|M|B\rangle=\sqrt{(1-\mu(A))(1-\mu(B))}\cos\beta$\\
2.   &  $\phi_j ={1 \over f_j} u_j$ \\
3.   &  $J_{ij} = \frac{|N_i \cap N_j|}{|N_i \cup N_j|},$\\
4.   &  $\sum_j {f_{ij} \over f_i} \phi_j$\\
5.   &  $\mathbf{M}_{21} \leftarrow \sum^{I_3}_{i=1} \mathbf{X}_{::i}^{\mathsf{T}}\mathbf{A} \mathbf{A}^{\mathsf{T}} \mathbf{X}_{::i}$\\
\hline
\bottomrule
\end{tabular}}
\end{small}
\end{center}
\vskip -0.1in
\caption{PV-DM: Example query equation (top row) and its top 5 nearest equations discovered model.}
\end{minipage}\hfill

\vspace*{.8cm}

\begin{minipage}{\textwidth}
\centering
\begin{center}
\begin{small}
{\renewcommand{\arraystretch}{1.4}
\begin{tabular}{ll}
\toprule
\hline
\multicolumn{2}{c}{Model: GloVe} \\
\midrule
\multicolumn{2}{c}{Query Equation: $\cos(\textbf{x},\textbf{y}) = \frac{\sum^n_{i=1}x_{i} \cdot y_{i}}{\sqrt{\sum^n_{i=1}x^2_{i} \cdot \sum^n_{i=1} y^2_{i}}}$}\\
\midrule
Rank. &  \multicolumn{1}{c}{\textbf{Top Equations}} \\
\hline
1.   &    $d_H^2(u,v)=\sum_{i=1}^V(\sqrt{u_i}-\sqrt{v_i})^2.$   \\
2.   &   $P[match(A,B)] = 1 - (1 - s^K)^M$    \\
3.   &    $s_n = \arg\min_{f \in V\backslash S_{n-1}}(\lambda \chi^2(f,V\backslash S_{n-1}) - (1-\lambda)\min_{s \in S_{n-1}} \chi^2(f,s))$   \\
4.   &    $P(t_{name}|s_{name})  =  \frac{P(s_{name}|t_{name}) *  P(t_{name})}{P(s_{name})}$   \\
5.   &    $Q' = \frac{1}{2m} \sum_{i}\sum_{j} \left(A_{ij} - \frac{k_i k_j}{2m} \right) s_i s_j,$   \\
\hline
\bottomrule
\end{tabular}}
\end{small}
\end{center}
\vskip -0.1in
\caption{GloVe: Example query equation (top row) and its 5 nearest equations discovered with this model.}
\end{minipage}
\end{table}
\vspace*{2cm}

\begin{table}[t]
\begin{minipage}{\textwidth}
\centering
\begin{center}
\begin{small}
{\renewcommand{\arraystretch}{1.4}
\begin{tabular}[b]{ll}
\toprule
\hline
\multicolumn{2}{c}{Model: EqEmb} \\
\midrule
\multicolumn{2}{c}{Query Equation: $\cos(\textbf{x},\textbf{y}) = \frac{\sum^n_{i=1}x_{i} \cdot y_{i}}{\sqrt{\sum^n_{i=1}x^2_{i} \cdot \sum^n_{i=1} y^2_{i}}}$}\\
\midrule
Rank. &  \multicolumn{1}{c}{\textbf{Top Equations}} \\
\hline
1.   &    $similarity(v_p, v) = \frac{\sum_i^n v_{p_i} v_i}{\sqrt{\sum_i^n v_{p_i}^2}  \sqrt{\vphantom{\sum_i^n v_{p_i}^2} \sum_i^n v_i^2}}$ \\
2.   &    $0\le f_c=\min({\mu(A)+\mu(B) \over 2} - \mu(A\ {\rm and} \ B),\mu(A\ {\rm and}\ B)-\mu(A)\mu(B))$\\
3.   &    $o = \sum_i p_i c_i. $ \\
4.   &    $\ell_{QQ}({q}, {q}', {q}'') = \big[\gamma - {S}_{QQ}({q}, {q}') + {S}_{QQ}({q}, {q}'') \big]_+\,.$\\
5.   &    $\mathrm{Sim}_{\gamma} ( P, Q) = \frac{\sum_{i = 1}^\ell p_i^\gamma q_i^\gamma} {\sqrt{\sum_{i = 1}^\ell p_i^{2\gamma}} \sqrt{\sum_{i = 1}^\ell q_i^{2\gamma}}}$\\
\hline
\bottomrule
\end{tabular}}
\end{small}
\end{center}
\vskip -0.1in
\caption{EqEmb: Example query equation (top row) and its 5 nearest equations discovered with this model.}
\end{minipage}\hfill
\end{table}

\vspace*{16cm}

\section{Additional Examples of Query Equations}
\label{sec:example_query_eq}
In this appendix we give an additional examples of query equations and their 5 nearest equations and words.

\begin{table}[h]
\begin{minipage}{\textwidth}
\begin{center}
\begin{small}
{\renewcommand{\arraystretch}{1.6}
\begin{tabular}{clcl}
\toprule
\hline
\multicolumn{4}{c}{$h_s^t=o_t \cdot c_t$}\\
\midrule
\multicolumn{2}{c@{\quad}}{Top Equations} & \multicolumn{2}{c@{\quad}}{Top Words} \\\cmidrule(r){1-2}\cmidrule(l){3-4}
1.  & $c_t=f_t \cdot c_{t-1}+i_t \cdot l_t$  &1.  & lstm \\
2.  &  $y=g(W^ox_{\rho}+b^o)$   & 2.  & layer \\
3.  &  $h_t=f(e^T \cdot W \cdot e + v \cdot e) $   &3.  & trained \\
4.  &  $h_t=o_t \odot c_t, $   & 4.  & generates\\
5.  &  $\tilde{c}_t = tanh(W_cE[y_{t-1}]+U_ch_{t-1}+A_c\varphi_t(V)+b_c). $ & 5.  & rnn \\
\bottomrule
\end{tabular}}
\end{small}
\end{center}
\caption{Example query equation (top row) and its 5 nearest equations (left) and words (right) discovered using EqEmb. All similar equations relate to neural network layers such as the query equation.}
\label{tab:ir_example2}
\end{minipage}\hfill
\end{table}

\appendix
\end{document}